\documentclass[journal]{IEEEtran}
\IEEEoverridecommandlockouts
\usepackage{cite}
\usepackage{booktabs}
\usepackage{amsmath,amssymb,amsfonts}
\usepackage{algorithmic}
\usepackage{graphicx}
\usepackage{textcomp}
\usepackage{amsmath,amssymb,amsfonts}
\usepackage{algorithmic}
\usepackage{array}
\usepackage{tabu}
\usepackage{tabularx}
\usepackage{svg}
\newcommand{\eatme}[1]{ }
\usepackage[inline]{enumitem}
\usepackage{float}
\usepackage{cite}
\usepackage{amsmath,amssymb,amsfonts}
\usepackage{algorithmic}
\usepackage{array}
\usepackage{tabu}
\usepackage{graphicx}
\usepackage{textcomp}
\usepackage{subcaption,siunitx,booktabs}
\usepackage{multirow}
\usepackage{ragged2e}
\usepackage{tabularx,tabulary}
\usepackage{adjustbox}

\renewcommand{\thetable}{\arabic{table}}   
\renewcommand{\thefigure}{\arabic{figure}}
\usepackage{xspace}
\usepackage{url} 
\usepackage{makecell}
\usepackage{fixltx2e}

\def\BibTeX{{\rm B\kern-.05em{\sc i\kern-.025em b}\kern-.08em
    T\kern-.1667em\lower.7ex\hbox{E}\kern-.125emX}}

\begin{document}

\title{Graph Attention Network for Lane-Wise and Topology-Invariant Intersection Traffic Simulation\\
}

\makeatletter
\newcommand{\linebreakand}{%
  \end{@IEEEauthorhalign}
  \hfill\mbox{}\par
  \mbox{}\hfill\begin{@IEEEauthorhalign}
}

\author{Nooshin Yousefzadeh, Rahul Sengupta, Yashaswi Karnati,  Anand Rangarajan, and Sanjay Ranka\\
\textit{Department of Computer and Information Science and Engineering} \\
\textit{University of Florida, Gainesville, FL, USA} \\
\{nooshinyousefzad, rahulseng, yashaswikarnati,anandr,sranka\}@ufl.edu}

\maketitle

\begin{abstract}

Traffic congestion has significant economic, environmental, and social ramifications. Intersection traffic flow dynamics are influenced by numerous factors. While microscopic traffic simulators are valuable tools, they are computationally intensive and challenging to calibrate. Moreover, existing machine-learning approaches struggle to provide lane-specific waveforms or adapt to intersection topology and traffic patterns. In this study, we propose two efficient and accurate "Digital Twin" models for intersections, leveraging Graph Attention Neural Networks (GAT). These attentional graph auto-encoder digital twins capture temporal, spatial, and contextual aspects of traffic within intersections, incorporating various influential factors such as high-resolution loop detector waveforms, signal state records, driving behaviors, and turning-movement counts. Trained on diverse counterfactual scenarios across multiple intersections, our models generalize well, enabling the estimation of detailed traffic waveforms for any intersection approach and exit lanes. Multi-scale error metrics demonstrate that our models perform comparably to microsimulations. The primary application of our study lies in traffic signal optimization, a pivotal area in transportation systems research. These lightweight digital twins can seamlessly integrate into corridor and network signal timing optimization frameworks. Furthermore, our study's applications extend to lane reconfiguration, driving behavior analysis, and facilitating informed decisions regarding intersection safety and efficiency enhancements. A promising avenue for future research involves extending this approach to urban freeway corridors and integrating it with measures of effectiveness metrics.

\end{abstract}

\begin{IEEEkeywords}
Traffic, Intersection, Waveform, Graph Neural Networks, Deep Learning, Graph Attention Networks, ATSPM
\end{IEEEkeywords}

\section{Introduction} 
Traffic congestion has a major impact on the economy, environment sustainability, public health, and overall life quality of societies. According to the Texas A\&M Transportation Institute's Urban Mobility Report (2021), congestion costs Americans nearly \$166 billion in 2019 in terms of wasted time and fuel costs \cite{MobilityDivision2022}. Improving traffic signal control can greatly help in mitigating traffic congestion \cite{guo2019urban}. 

 Automated Traffic Signal Performance Measures are a collection of analytics tools that convert high-resolution loop detector data and signal controller data into actionable performance measures \cite{876393:20300507}. ATSPM gives data-driven insights to traffic engineers to adjust signal timing parameters and help optimize traffic flow.


Traffic flow at an intersection is affected by a host of factors. This includes green time assigned to each phase in actuated intersections, which varies from cycle to cycle. Turning-movement counts (TMCs) represent the counts of vehicles based on the various turn movements (left, through, and right) the vehicles can make. Driving behaviors can vary based on local driving patterns, type of area, vehicle composition, weather, etc. Further, the geometry of the roadways and intersection topology also impact traffic flow. Moreover, traffic patterns may change hourly, daily, weekly, seasonally, or in response to an incident.



Macroscopic simulators cannot provide fine-grained space-time analysis of traffic behavior. Microscopic simulators such as ReTime \cite{alshayeb2023optimizing}, VISSIM \cite{Shah2017ApplicationOV}, and SUMO (Simulation of Urban MObility) \cite{876393:20293669}, on the other hand, are more realistic but computationally-expensive in their application to signal timing optimization since every change in traffic flow and behavior (due to varying times of day, days of week, weather conditions etc.) necessitates a whole slew of fresh simulation runs. Thus, there is a need for efficient and reusable means of modeling intersection dynamics.

Another issue with using microsimulation for signal plan optimization is the realistic generation of incoming traffic platoons. The readings captured at the stop-bar are highly correlated to the signal timing plan. Thus, it is also important to estimate the traffic flow sufficiently upstream of the intersection, before the incoming traffic comes within the proximity of the intersection, and thus gets affected by the vehicle queues and traffic light conditions.


In this study, we introduce graph-based deep learning \textit{Digital Twins} to an intersection. They are inductive (based on the attention mechanism), and generative (based on latent representation). This attention graph-based auto-encoder architecture leverages the recent advent of automated traffic signal performance measures (ATSPM) systems that provide loop detectors and signal timing information at the decisecond level for more accurate and realistic modeling of traffic flow dynamics. Our models are trained on a wide range of traffic flow scenarios across a 9-intersection corridor, with varying signal timing plans and driving behavior parameters. Further, our models can handle varying intersection topologies, with different numbers of incoming and outgoing lanes. Specifically, given the stop-bar detector readings (waveforms), we model the exit detector readings (waveforms), as well as the incoming traffic readings (waveforms) upstream of the intersection. The design of our models mainly relies on the \textit{Attention} mechanism used with the Graph Neural Networks (GNNs) architecture to make it maximally informative on unseen intersections and traffic scenarios. The models are trained on two diverse datasets based on real-world as well as random traffic scenarios, across a wider range of varied signal timing plans to show its robustness. 

The experimental results show the effectiveness of the self-attention module in the temporal decoding of stop-bar loop detector waveforms, and the effectiveness of attention in the aggregation mechanism used in the method of message propagation. The model performs well on both datasets, although its performance is slightly better on the Random-TMC dataset, it is because the range of variations in real traffic scenarios is restricted. Moreover, the marginal improvement at higher levels of aggregation is better, which implies the estimated waveforms tend to be more accurate and consistent with the overall trends and patterns when short-term fluctuations and variations are smoothed out. We also used some explanation tools to detect the most effective features while reconstructing the exit waveforms, as well as several distinct embeddings in the latent representations associated with each group of lanes at the direction of movement.

The main contributions of this work can be summarized as follows:

\begin{enumerate}
    \item We introduce graph auto-encoder digital twins employing Graph Attention Neural Networks (GAT) to model fine-grained, lane-wise traffic flow dynamics simultaneously as vehicles approach and exit any intersection. These twins leverage high-resolution loop detector data and signal timing information for accurate representation.
    
    \item In real-world scenarios, stop-bar detectors are predominantly deployed. Our digital twins rely exclusively on stop-bar detectors to estimate exit and inflow waveforms, ensuring practical applicability and feasibility.
    
    \item Our digital twins provide estimations tailored to current traffic conditions by integrating multiple influential factors, including signal timing plans, driving behavior parameters, and turning-movement counts, into the graph model input.
    
    \item Employing attention mechanisms, our digital twins capture temporal, spatial, and contextual aspects of traffic flow gleaned from diverse intersections with varying topologies. During batch inference, the same trained model efficiently estimates exit and inflow waveforms under unseen traffic conditions at arbitrary intersections within seconds, offering superior efficiency compared to running parallel microsimulation instances and parsing their logs.
    
    \item Our digital twins enable rapid prediction of the impact of individual influential factors on lane-wise platoons approaching and traversing intersections, outperforming microscopic simulators in computational efficiency. Our approaches entail $O(1)$ sequential computation and can be fully parallelized with cost-effective GPU computation.
    
\end{enumerate}

Compared to the previous work by Karnati et. al. \cite{karnati2021subcycle}, this approach exploits graph neural networks instead of encoder-decoder architecture to preserve the topology of the intersection in the computation. Thus, the proposed model is able to expand the waveform estimation from a single intersection with two lanes on each of the four approaches to a general real-life intersection with a varying number of lanes per approach. Using this approach, our proposed models are able to simultaneously provide fine-grained lane-wise estimation of traffic flow dynamics at all inbound/outbound directions of an intersection with arbitrary topology. Thus, the same model can be trained to be applied to different intersections.

We also address two limitations of this study. First, the models use only stop-bar induction loop detectors to make predictions. This is based on our experience collecting real-world field data, where only stop-bar data is usually captured and stored. Second, the phase order of the signal timing plans, along with red and yellow times has been kept fixed. The green splits, however, do vary based on actuated Ring-and-Barrier control. Usually, the phase order is determined by the local traffic authority based on concerns such as safety and throughput, and we have used field settings. We have however varied other signal parameters such as cycle length and barrier times, within reasonable limits.



\begin{figure}
        \centering
        \captionsetup{justification=raggedright,singlelinecheck=false}
        \includegraphics[height = 2in,width=1\columnwidth]{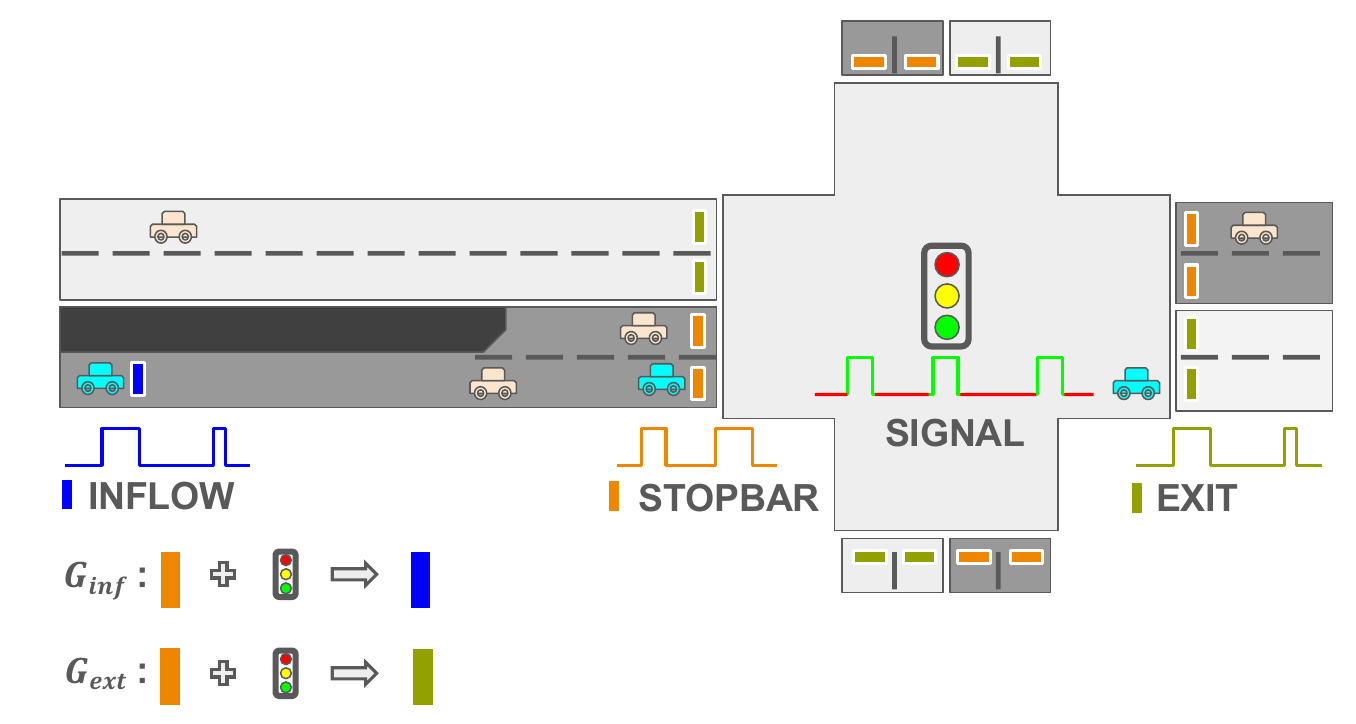}
        \renewcommand{\figurename}{Figure}
        \captionsetup{size=footnotesize}
\footnotesize
        \caption{\textbf{The physical location of Stop-bar and virtual location of Exit and Inflow loop detectors in the simulation of an intersection.} We simulate the ATSPM time series waveform within 8-phase standard NEMA phasing intersections. We train two digital twins that can estimate downstream exit waveforms of every outflow lane of all directions ($G_{ext}$) or upstream inflow waveforms of every inflow lane in all directions ($G_{inf}$) simultaneously for an intersection with arbitrary topology and characteristics.}
        \label{fig:sample_1_ins}
    \end{figure}

The rest of the paper is outlined as follows:
In Section \ref{prelim}, we briefly introduce several common terms used in deep learning. 
We define the problem formally in Section \ref{probdefn}. 
 In Section \ref{proposedmodels}, we present different models that we develop to relate various observable and unobservable quantities. Section \ref{datagen} describes how we preprocess raw data and generate synthetic datasets from real-world controller logs data using SUMO. Experimental results are provided in Section \ref{expresults}. Section \ref{sec:related} presents the related work of different techniques used for traffic state estimation. We finally conclude in Section \ref{conclusion}.


\section{Problem Definition}
\label{probdefn}
The datasets used to train our digital twins are constructed through 40,000 simulation hours. Each traffic simulation scenario runs for one hour and consists of a base map containing static components that remain unchanged (e.g., road segments, lane design) and dynamic components that vary in state or position across different scenarios (e.g., traffic signals, vehicles). Simulations are executed with random settings of configurable parameters serving as input parameters for our digital twins.

During simulation execution, log files are generated, which are subsequently parsed to create a "Simulation Record," documenting the details of each simulation run. In the following sections, we introduce terminology that formalizes the problems we aim to address. For a comprehensive understanding of the notations used in this study, detailed descriptions are provided in Table \ref{table:notation}.

\noindent\textbf{Definition 1. Simulation Record:} An extraction of data from log files parsed after execution of one single simulation run. A Simulation Record contains the information required to construct our training datasets. A Simulation Record can be defined as a tuple $s = {(j, sig_{j,w}, tmc_{j}, drv_{j}, stp_{j,w}, ext_{j,w}, inf_{j,w})}$, where the terms refer to the intersection ID, signal waveform, turn-movement counts vector, driver behavior vector, stop-bar loop detector waveform, exit loop detector waveform, and inflow loop detector waveform respectively.

\noindent\textbf{Definition 2. Exit Simulation Graph:} This term refers to a single graph data item within the graph dataset utilized for training and evaluating our Exit waveform estimator digital twin. Each Exit Simulation Graph adheres to a standardized structure, representing a specific traffic scenario occurring at one of nine intersections contributing to the construction of the training data. Given a \textit{Simulation Record} $s$, for an intersection $j$, an Exit Simulation Graph $g_{s,j}=(V, E, X, Z)$ can be constructed as a connected single-layer directed bipartite graph, with a relevant common structured adjacency list as described in Subsection \ref{graphrep}. The node set $V \in \mathbb{R}^{1 \times N}$ represents lanes situated one hop away upstream and downstream with respect to the intersection. The edge set $E \in \mathbb{R}^{2 \times M}$ depicts permissible turning-movements \footnote{The permissible turning-movements are assumed to encompass left-turns, through movements, and right-turns. We do not consider u-turns with the assumption that they coincide with left-turns} between any pair of stop-bar and exit waveforms. Each incoming lane (upstream) is equipped with its associated stop-pbar detector waveform, while each outgoing lane (downstream) is equipped with its associated exit detector. Consequently, the node feature matrix $X \in \mathbb{R}^{N \times w}$ comprises all loop detector waveform time series associated with either a stop-bar or an exit loop detector. The edge features $Z = [sig_{j,w}, tmc_{j}, drv_{j}]$ are a concatenation of the signal timing plan time series, a vector representing turning-movement counts, and a vector representing driving behavior parameters.

\noindent\textbf{Definition 3. Inflow Simulation Graph:} This term denotes a singular data item within the graph dataset used for training and assessing our Inflow waveform estimator digital twin. Each Inflow Simulation Graph conforms to a standardized structure, representing a distinct traffic scenario occurring at one of nine intersections contributing to the training data. The key distinction between an Inflow Simulation Graph and an Exit Simulation Graph lies in their construction. The former is fashioned as a connected multi-layer directed bipartite graph. Each layer of this graph replicates the same set of nodes, representing a single road segment intersecting an intersection. The edge set $E \in \mathbb{R}^{2 \times M'}$ establishes a full connection between each pair of stop-bar and inflow waveforms, facilitating the modeling of all conceivable lane-changing movements between any two lanes. Additionally, supplementary edge items, termed pillar edges, are introduced to link each node to its counterpart in any other layer. As a result, the node feature matrix $X \in \mathbb{R}^{N' \times w}$ encompasses all loop detector waveform time series associated with either a stop-bar or an inflow loop detector. The edge features $Z = [sig_{j,w}, tmc_{j,l}, drv_{j}]$ consist of a concatenation of the signal timing plan time series, a vector representing turning-movement counts, and a vector representing driving behavior parameters. Notably, the turning-movement counts vary from one layer to another, offering specificity to the traffic pattern of each approach within an intersection.

\noindent\textbf{Problem 1. Exit Waveform Estimation:} Our objective is to simultaneously simulate traffic waveforms exiting from every lane of any direction within an intersection. We aim for this estimation to be applicable to any arbitrary topology of an intersection while remaining specific to the current state of signal timing plans, driving behaviors, and turning-movement counts. To address this challenge, we introduce the $G_{ext}$ digital twin. Leveraging numerous Simulation Records from various intersections, we construct a comprehensive dataset of Exit Simulation Graphs $g_{s,j} = (V, E, X, Z)$, where the node feature matrix $X$ is specifically masked for exit waveforms. Our goal is to develop a graph-based auto-encoder $G_{ext}: (A, stp, sig, tmc, drv) \rightarrow ext$, incorporating a suitable message propagation scheme. This model can effectively learn from the constructed graphs to impute missing exit waveforms. 

\noindent\textbf{Problem 2. Inflow Waveform Estimation:} Our objective is to simultaneously simulate traffic waveforms approaching from every lane of any direction within an intersection. We aim for this estimation to be applicable to any arbitrary topology of an intersection while remaining specific to the current state of signal timing plans, driving behaviors, and turning-movement counts. To address this challenge, we introduce the $G_{inf}$ digital twin. Leveraging numerous Simulation Records from various intersections, we construct a comprehensive dataset of Inflow Simulation Graphs $g'_{s,j} = (V', E', X', Z')$, where the node feature matrix $X$ is specifically masked for inflow waveforms. Our goal is to develop a graph-based auto-encoder $G_{inf}: (A', stp, sig, tmc', drv) \rightarrow inf$, incorporating a suitable message propagation scheme. This model can effectively learn from the constructed graphs to impute missing inflow waveforms.

\subsection{Common Structure for Graph Adjacencies}
\label{graphrep}

The graph-based digital twins undergo training using a plethora of graph data points, all of which must adhere to a standardized structure. We employ a uniform structure for constructing Exit Simulation Graph datasets, characterized by a single-layer graph representation, and another structure for constructing Inflow Simulation Graph datasets, featuring a multi-layered graph representation. In this study, we focus on a standard NEMA 4-way intersection with north-bound, south-bound, east-bound, and west-bound approaches. However, the number of constituent lanes may vary across intersections. To ensure consistent lane connectivity in the input graph, we standardize the presentation of adjacency information. In cases where an approach or multiple approaches are missing, dummy lanes with zero traffic are introduced to maintain uniformity. 

The common adjacency list for the Exit Simulation Graph for a template intersection represents a single-layer bipartite graph, connecting all possible incoming lanes (featured with relevant stop-bar loop detector waveform) to all possible outgoing lanes (featured with relevant exit waveform). This standardized topology encompasses the maximal set of incoming-outgoing connections that any intersection within the corridor may have along any approach. For instance, if the highest number of left-turning lanes along the East-bound approach (across all intersections in the corridor) is 2, then this number is chosen for East-bound left-turns. Connections that are unavailable at specific intersections are denoted as "dummy" nodes, where both the stop-bar and exit waveforms are consistently zero, indicating no traffic actuation between those two lanes.

We observed, that the total number of 22 connections suffice to capture all possible connectivity between 22 incoming lanes and 11 outgoing lanes across all contributing intersections in this study. Therefore, the common form of adjacency matrix $A \in \mathbb{R}^{22 \times 11}$ for this single-layer graph can be constructed.

The common supra adjacency list for the Inflow Simulation Graph for a template intersection constitutes a multi-layer graph \cite{yousefzadeh2023comprehensive}, where each layer constructs a bipartite graph connecting lanes with stop-bar detectors to lanes with inflow detectors at each approach. These layers encompass the lanes situated one hop away from the intersection (housing stop-bar detectors) and the lanes feeding into them (housing inflow detectors).

We observe that there are at most 6 1-hop lanes (at the stop-bar) and at most 3 inflow lanes feeding into them to cover all topologies of contributing intersections in this study. At each approach, there are 18 potential combinations that a vehicle can traverse as it crosses one of the inflow detectors and subsequently one of the stop-bar detectors before exiting the intersection. Furthermore, each stop-bar and inflow detector within one approach is linked to its counterparts across other approaches within the same intersection, signifying their commonality that counts for 108 pillar edges. With 6 1-hop lanes and 3 feeding lanes, a total of 9 lanes of interest exist in each of the 4 approaches, the common form of adjacency matrix $A' \in \mathbb{R}^{24 \times 12}$ for this multi-layered graph can be constructed.

\begin{table}[ht]
    \begin{adjustbox}{width=\columnwidth,center}
        \begin{tabulary}{1.0\textwidth}{|l|l|l|l|l|l|}
            \hline
            \textbf{Notation}&\textbf{Description}  &\textbf{Aggr}  &\textbf{Size}  &\textbf{Type}  \\ 
            \hline
            $j$&Intersection&-&1x9&String\\\hline
            $l$&Lane segment&-&-&String\\\hline
            $stp$&Waveform at stop-bar detector  &5 sec  &1x80  &Integer, 0-8 \\\hline
            $ext$&Waveform at the exit of the intersection  &5 sec  &1x80  &Integer, 0-8 \\\hline
            $inf$&Waveform upstream the intersection  &5 sec  &1x80  &Integer, 0-8 \\\hline
            $sig$&Signal timing state information  &5 sec  &8x80  &Binary \\\hline
            $drv$& Driving behavior parameters  &2400 sec  &1x9 &Float, 0-30\\\hline
            $tmc$& Turning-movement counts ratio &2400 sec  &4x3  &Float, 0-1 \\\hline
            $w$& Size of the observation window&-  &1  &Integer, 0-150 \\\hline
            $w'$& Size of the prediction window&-  &1  &Integer, 0-150 \\\hline
            $s$& Single simulation recorded by SUMO &5 sec&-&Multiple \\\hline
            $G_{ext}$& Digital twin for Exit experiment&-&-&-\\\hline
            $G_{inf}$& Digital twin for Inflow experiment&-&-&-\\\hline
            $A$& Common lane-connectivity for $G_{ext}$&-&22x11&Binary\\\hline
            $A'$& Common lane-connectivity for $G_{inf}$&-&(4x6)x(4x3)&Binary\\\hline
            $N$&Number of lanes in $A$  &- &1x33  &Integer \\\hline
            $M$& Number of lane connections in $A$ &- &1x22  &Integer \\\hline
            $N'$&Number of lanes in $A'$  &- &4x9  &Integer \\\hline
            $M'$& Number of lane connections in $A'$ &- &4x18+4x27  &Integer \\
            \hline
        \end{tabulary}
    \end{adjustbox}
\renewcommand{\tablename}{Table}
\caption{Summary of the notations and their definitions.}
\label{table:notation} 
\end{table}

\section{Preliminaries}
\label{prelim}

\subsection{Deep Learning} Deep Learning \cite{Goodfellow-et-al-2016} is a sub-field of Artificial Intelligence (AI) that uses neural networks with multiple layers (known as \textit{deep neural networks}) in order to extract a hierarchical representation of features from raw data, and progressively learn complex representations while performing various machine learning tasks. 


\subsection{Attention Mechanism} Attention Mechanism \cite{vaswani2017attention} is a machine learning technique that has its roots in human cognitive processes, allowing models to have a dynamic focus on relevant parts of input data while making predictions. Attention allows the model to adaptively allocate varying degrees of focus to different parts of the input sequence, to model long-range dependencies between its distant elements. 


The variants of attention mechanisms are proposed to compute the attention weights, in order to capture the complexity of the relationships within the input sequence. A common type of attention mechanism employed in self-attention mechanisms is \textit{scaled dot-product attention}, in which attention weights are computed using the dot product of query and key, followed by scaling and Softmax normalization.

\subsection{Graph neural networks} Graph Neural Networks (GNNs) \cite{gnn1} are designed to perform various machine learning tasks by leveraging the inherent structural information in the input graph in the form of relationships (i.e. edges) that inter-connect entities (i.e. nodes). GNNs have been shown to be effective in many fields \cite{wu2020comprehensive} such as social network analysis, recommendation systems, bioinformatics, and knowledge graph completion, etc.

A common thread across the variants of GNN architectures is the graph \textit{message passing} learning paradigm that allows nodes (and/or edges) to update their representations iteratively by exchanging information with their immediate neighbors. This enables the model to capture hierarchical and recursive patterns in input graph(s). The node representations, called node embedding, after the $k^{th}$ GNN layer are computed as below, where $\sigma(.)$ denotes an activation function, such as the $ReLU$ (Rectified Linear Unit), $N(i)$ denotes the set of neighboring nodes to the node $i$, $W_k$ is the weight matrix associated to the $k^{th}$-order neighborhood, and $B_k$ is a weight vector associated with the node's own representation at $k^{th}$ step of propagation.

\[\
h_i^k = \sigma \Biggl( \sum_{j\in \mathcal{N}i)} \frac{h_j^{k-1}}{|\mathcal{N}(i)}|+B^k h_i^{k-1} \Biggl)
\]

Graph Convolutional Networks (GCNs) are a variant of GNNs and have laid the foundation for many subsequent Graph Neural Network architectures with applications in various tasks \cite{kipf2016semi}. GCNs adapt the concept of Convolutional Neural Networks (CNNs) to graph data by aggregating information from neighboring nodes for effective compression of both local and global structural patterns. 
The message propagation paradigm in GCNs is formulated by the following steps:
\begin{enumerate}
\item Initialization: Assigning node feature vectors and topological information.
\item Message Aggregation: Weighted sum of neighboring node features where the weighted are determined by graph structure.
\item Normalization: Prevent vanishing gradient problem by balancing the influence of different nodes.
\item Pooling: Combine normalized messages with the node’s representation.
\item Iteration: Repeating the above steps to refine node representation by considering information from a farther distance.
\end{enumerate}

The normalization term denoted by $c_{ij}$ is defined by the square root of the degrees of the nodes, the node representation after the $k^{th}$ layer as follows: 
\[\
h_i^k = \sigma \Biggl( \sum_{j\in \mathcal{N}(i)} \frac{1}{c_{ij}}W_k h_i^{k-1} \Biggl)
\]
With $k$-layers of GCN, node representations after $k^{th}$ layer are generated by propagating information at most $k$-steps away so as to encompass a viewpoint of all neighbors up to $k$-hops.

\subsection{Graph Attention Networks} Graph Attention Networks (GATs) are another popular class of GNNs that allow nodes to adaptively capture both local and global patterns, employing attention mechanisms while processing graph-structured data and capturing relationships between nodes \cite{ velivckovic2017graph}. GATs also introduce multiple attention heads to capture diverse relationships and interactions within the input graph(s).

At each GAT layer, the self-attention mechanisms compute, normalize, and assign attention coefficients $a_{ij}=softamx_j(e_{ij})$ to neighboring nodes as a measure of compatibility with the central node, for a contextually informative node representation after the $k^{th}$ layer, where $\parallel$ represents concatenation in the aggregation process of the multi-head graph attentional layer.

\[\
h_i^k = \parallel_{k=1}^K \sigma \Biggl( \sum_{j\in \mathcal{N}(i)} a_{ij}^k W_k h_j \Biggl)
\]

 Specifically in this paper, we choose to perform an aggregation process on the prediction layer by simply \textit{averaging} the output of the final non-linearity function, resulting in the following node representation:
\[\
h_i^k = \sigma \Biggl( \frac{1}{K}\sum_{k=1}^K \sum_{j\in \mathcal{N}(i)} a_{ij}^k W_k h_j \Biggl)
\]

\subsection{Graph Sampling Networks} GraphSAGE (SAmple and aggreGatE) Networks are another class of GNNs that is proposed for learning large and evolving networks, exploiting both node features and dynamic topological structure of each node's neighborhood \cite{hamilton2017inductive}. Unlike other graph learning methods like GCNs, GraphSAGE can integrate node features and generate embedding for new (unseen) nodes during training.  The node representation after the $k^{th}$ layer with element-wise mean aggregation, is as follows:
\[
h_i^k= \sigma(W_1^k h_i^{k-1}+W_2^k  mean_{j \in \mathcal{N}(i)} h_j^{k-1})
\]
The aggregation mechanism in GAT is attention-based and incorporates both node and edge features in its message propagation scheme.

\section{Proposed Models}
\label{proposedmodels}

This section introduces deep-learning approaches to build proposed digital twins.\footnote{Our implementation is available at https://github.com/NSH2022/Digital-Twin-Waveform.} Both of our digital twins use the concept of \textit{Attention} as a fundamental building block in their architectures. \textbf{Stop-bar} detectors are located at the stop-bars of various lanes at the intersection, whereas \textbf{Exit} loop detectors are at the start of each outflow lane, and \textbf{Inflow} loop detectors at 500 meters upstream (or at the mid-point between two intersections if they are less 750 meters apart) from each intersection.
Figure. \ref{fig:sample_1_ins} shows the location of each type of loop detector in the base map of the microscopic simulator.

\subsection{Exit Waveform Reconstruction}
\label{propEXIT}
Proposed $G_{ext}$ digital twin is an attention-based graph auto-encoder \cite{kipf2016variational} using a common structure of connectivity matrix (cf. Subsection \ref{graphrep}). It can be applied to intersections with any arbitrary topology for a fine (lane-wise) estimation of traffic flow exiting the intersection. This model couples a self-attention layer with a GATConv message-passing layer to decode temporal dependencies within individual input stop-bar time series waveforms and spatial dependencies between lanes within and across intersections. For simplicity, we used one single GATConv encoder layer in our model, although more GAT layers can be used in practice to obtain better results. 

We solve \textit{Problem 1.} using a graph-based auto-encoder that estimates exit waveforms $ext$ directly from input traffic time series waveforms $stp$. During input ($X$) both $stp$ and $ext$ are concatenated, but the $ext$ portion is masked. During the reconstruction, the same input ($\hat{X}$) with $stp$ and $ext$ is presented as the target, but unmasked. Thus, the model is expected to reconstruct the masked portion i.e. $ext$ at the output.
The model is composed of three components: self-attention, encoder, and decoder. The self-attention component is implemented using a masked scaled dot product attention mechanism
Masking attention in the self-attention layer ensures that prediction for position $i$ depends only on known outputs at positions less than $i$, while the residual connection prevents the co-adaptation of neurons by randomly dropping out some input variables. The output of self-attention layer $\hat{X}$ of size $ (N \times w)$ will be used to initialize node features for the encoder. The auto-encoder is composed of a GATConv \cite{velivckovic2017graph} message passing layer following a single fully connected linear layer parameterized by weight matrix $B \in \mathbb{R}^{w \times N}$ and activated by a ReLU function to generate and propagate messages (hidden representation of the states of inflow lanes) to any other connected node (outflow lanes) in the intersection network. In the formulation of this message passing scheme, the edge feature matrix of size $(M \times 29 \times w)$ is used, where 29 is the concatenation of various $tmc$, $drv$, and $sig$ variables. First, a learnable linear transformation generates a new set of node features with potentially different cardinality $X' \in \mathbb{R}^{N \times z}$ parameterized by a weight matrix $W \in \mathbb{R}^{N\times w}$, where $w$ and $z$ denote window size and hidden size of the model, respectively. Then a shared \textit{masked} attention mechanism computes attention coefficients while injecting graph structure information in the formulation. The attention mechanism in GATConv, itself is a SINGLE-layer feed-forward neural network, parameterized by a weight vector $a \in \mathbb{R}^{2N \times w}$ with ReLU non-linearity.
\[a_{ij}=ReLU(a^T[Wh_i\| Wh_j])\]
By using this technique, GATConv implicitly specifies and integrates the attention coefficients denoting the importance of each inflow lane $j$'s stop-bar time series waveform to each outflow lane $i$'s exit detector's time series without complicated matrix inverse operation or the need to know the graph structure upfront.

\[
\alpha_{ij}=Softamx_j(e_{ij}=\frac{exp(e_{ij})}{\sum_{k\in N_i}exp(e_{ik}))}
\]

Where $N_i$ is some lanes connected to lane $i$ in the intersection's graph representation. Finally, a decoder composed of a fully connected layer activated by a ReLU function to be applied on the output of the encoder to DIRECTLY impute missing (unobserved) time series waveforms for exit loop detectors $ext$ in reconstructed input feature matrix $\hat{X} \in \mathbb{R}^{N \times w}$.

\begin{figure*}
        \centering
        \captionsetup{justification=RaggedRight, size=footnotesize}
        \includegraphics[height =2.3 in,width=1\textwidth]{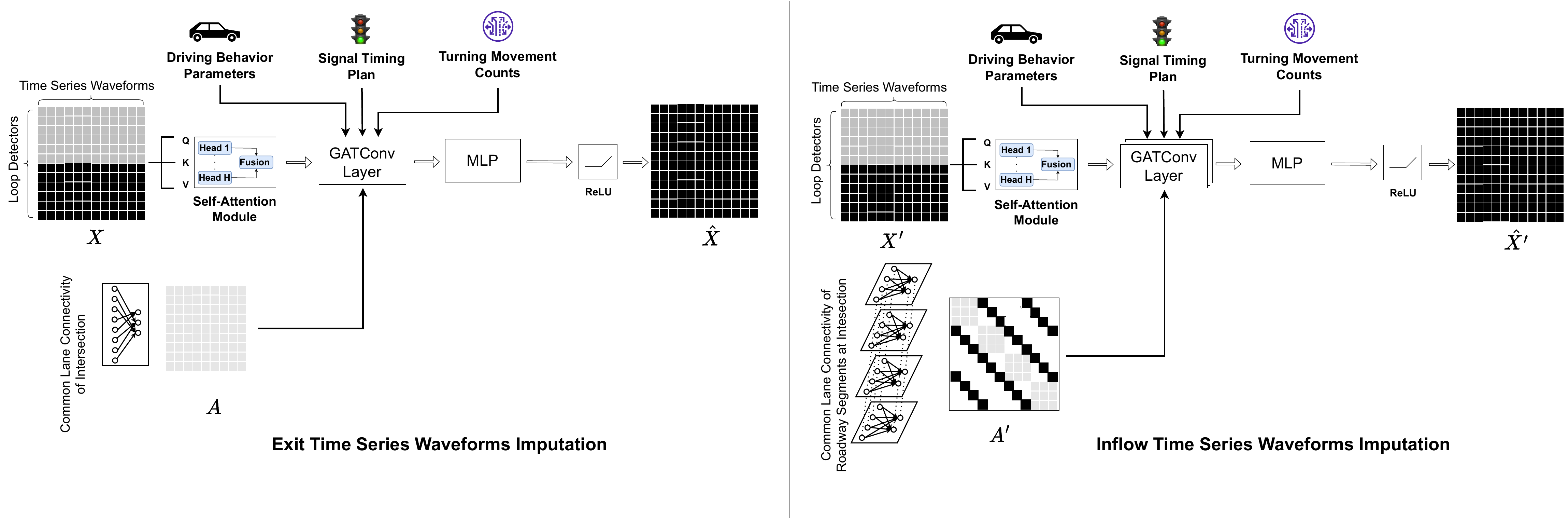}
        \renewcommand{\figurename}{Figure}
        \footnotesize
        \caption{\textbf{Overview of a proposed digital twin as applied to a single graph of traffic simulation for standard NEMA phasing intersection.} The right subplot shows architecture of $\mathbf{G_{ext}}$ model. 
        Masked node features $X$ are fed into a self-attention module for temporal encoding and then into a single GAT layer for spatial encoding while incorporating all effective factors specific to this specific traffic scenario. The decoder imputes missing (masked) time series waveforms for the exit loop detectors in the reconstructed feature matrix $\hat{X}$. The left subplot shows the architecture of $\mathbf{G_{inf}}$ model. Here the input of the model is multi-layered graph data and the imputed entities in the reconstructed feature matrix $\hat{X}$ are time series waveforms for the inflow loop detectors.}
        \label{fig:overview}
    \end{figure*}

\subsection{Inflow Waveform Reconstruction}
\label{propINF}
Proposed $G_{inf}$ digital twin uses the same graph attentional auto-encoder architecture as $G_{ext}$ using a different (multi-layer) common structure of connectivity matrix (cf. Subsection \ref{graphrep}), hence is still applicable to unseen intersections with any arbitrary topology for a fine (lane-wise), accurate and efficient estimation of inflow traffic flow upstream the intersection. 
Similarly, we solve \textit{Problem 2.} using a non-parametric graph-based auto-encoder with an imputation mechanism that works as an end-to-end solution to inflow waveforms reconstruction $inf$ directly from input traffic time series waveforms $stp$ as shown in Figure. \ref{fig:overview}-right subplot. In this model, the two layers of GATConv \cite{velivckovic2017graph} message passing layer following a single fully connected linear layer parameterized by weight matrix $W' \in \mathbb{R}^{w \times N'}$ and activated by a ReLU function to generate and propagate messages between any $stp$ node to any other connected $inf$ node in road segments around an intersection network. 
Attention coefficients generated by GATConv imply the importance of each stop-bar detector's time series waveform in the estimation of a particular inflow detector's time series located at the same road segment (layer of the graph). In this setting, the decoder is able to INVERSELY reconstruct missing (unobserved) time series waveforms for inflow loop detectors $inf$ in reconstructed input feature matrix $\hat{X'} \in \mathbb{R}^{N' \times w}$.

\section{Data Generation}
\label{datagen}

In this section, we discuss how we generated our exhaustive dataset using the SUMO simulation framework based on real-world parameters. The dataset is based on over 400,000 hours of simulation data of a real-world 9-intersection urban corridor in a large metropolitan region in the United States of America.
The traffic corridor used as the base map for simulation consists of 9 intersections with varying numbers of lanes at each approach. The intersections consist of four
through/right movements and four left-turn movements, one of each for the four approaches. The exception to this is the left-most (i.e. eastern-most) intersection, which is a T-intersection with only 3 approaches. Within the simulation, induction loop detectors have been installed at the stop-bars of the intersections. Further, advance loop detectors have also been placed upstream (usually around 100 meters away from the intersection). Further, "exit" detectors have been placed at the start of the outflow lanes at the intersections.

In order to generate a large comprehensive dataset, that contains a wide range of traffic behaviors, it is important to generate diverse traffic flows. Real-world recorded flows, based on sensing technologies such as induction loop detectors (ATSPM), as well as sparse probe trajectory GPS data (from vendors such as WEJO), are useful. ATSPM consists of fixed sensors that can capture nearly all the vehicles that traverse them, whereas sparse probe data usually has a penetration rate of 1\%-5\%. However, sparse probe data can give a representative view of the routes of vehicles plying on the corridor. Using the two data sources, it is possible to generate an approximate Origin-Destination probability matrix. This tells us the likelihood that a vehicle starting at a certain origin will go to which valid destination. We use SUMO's tool od2trips for O-D matrices for different hours of the day (especially between 6 am and 8 pm), to generate large route files.

However, since an important goal of the trained models is to be able to work for counter-actual traffic scenarios, we need physically possible but not non-observed flows. Towards this end, we generate random vehicle flows, which vary between 0 to two times the average real-world flows. These vehicles travel on random routes, giving rise to feasible yet unobserved turning-movement counts at the various intersections. These scenarios may lead to partial or network-wide congestion due to cross-blocking, insufficient green times, etc.

Since one of the uses of our digital twins is to aid signal timing optimization, it is important to vary the signal plans across the corridor within reasonable limits. We assume the intersections making up the corridor have Ring-and-Barrier operation with a common cycle length. For each simulation, a random common cycle length is chosen between 120 seconds to 240 seconds. Based on the selected common cycle length, random offsets are chosen for each intersection separately. Then for each intersection, a random barrier time is chosen, that separates the co-ordinated phases and the non-co-ordinated phases. The minimum/maximum green times, yellow times, and red times for the intersections are based on field values obtained from signal timing sheets. Driving behavior parameters are varied to ensure the models trained can operate on a wide range of behaviors. These include vehicle characteristics (such as acceleration, deceleration, and emergency braking), safety parameters (such as minimum gap between vehicles, and headway), and lane-changing parameters.

The dataset was generated using HiperGator supercomputing resources using parallel computing libraries such as multiprocessing.
\subsection{Data sets} 
Our dataset consists of around 400,000 exemplars across 9 intersections based on hour-long simulations. Half of them are based on inferred real-world arrival-departure routes of vehicles (OD matrices) to simulate traffic scenarios and the remaining on random routes. This diversity in data ensures our trained models not only learn on real-world traffic scenarios (based on route files using ATSPM and WEJO data) but also counterfactual ones (based on randomly generated route files). 
\begin{itemize}
\item Real-TMC Dataset: This dataset based on real-world routes is used to construct two graph datasets. To train $G_{ext}$ for the Exit experiment, each graph data represents an intersection with a common topology including 33 nodes (i.e., up to 22 inflow lanes and 11 outflow lanes) and 22 edges. To train $G_{inf}$ for the Inflow experiment, each tile layer of graph data represents a roadway segment of an up to four-way intersection with a common topology including 33 nodes with the same set of 9 nodes at each layer (i.e., up to 3 Inflow lane segments and 6 connected lanes segments at each direction) and 180 edges out of which 108 edges are pillars connecting every node in a layer and its counterpart in other layers). 
\item Random-TMC Dataset: This dataset based on randomized routes is used to construct two graph detests with the same above-described structures.
\end{itemize}

A complete list of varying parameters and the range of variation during the generation of traffic simulation records are listed in Supplementary Table 1. The randomly generated route files are varied in two important aspects:
\begin{itemize}
  \item The origin-destination probability matrices are randomly generated instead of being inferred from WEJO data. This gives rise to unseen turn-movement counts.

  \item The number of vehicles generated at a particular origin varied between 0 and two times the expected vehicle flows seen at that intersection. This gives rise to unseen vehicle demand patterns.
\end{itemize}

\section{Experimental Results}
\label{expresults}
In this section, we compare the performance of proposed digital twin models on our benchmark datasets. This comparison evaluates our digital twins with respect to the effectiveness of the lanes' hidden representations encoded from the different functions of effective factors provided by different message-passing formulations. We used the following Python packages e.g., Lightning, MLflow, and PyTorch Geometric. 
All models are optimized using the mean squared error (squared L2 norm) loss metric and converged under the same early stopping conditions. We further analyze and report the mean absolute error (MAE) in terms of \textit{vehicle per bucket} aggregated at coarser resolutions (10s, 15s, 20s, etc.). We used the learning rate of 0.001 with Adam's optimizer. 


We use different variants of our proposed digital twins by alternatively using nodes and edges to store the exemplar attributes. We also vary the encoder unit with other popular methods of message passing propagation i.e., SAGEConv \cite{hamilton2017inductive} (further referred to as \textbf{SAGEConv-EXT}), as well as GCNConv \cite{kipf2016semi} (further referred to as \textbf{GCNConv-EXT}). Another variant is constructed by ablation of the self-attention module used for temporal decoding of waveforms, (further referred to as \textbf{GATConv-Ablated}). We use a single graph convolution layer (except for \textbf{GATConv-INF} with two layers) for all models and 2 attention heads for GATConv models. Table. \ref{table:modelpar} is a comparison of the learning performance of proposed models and their variants in terms of the number of trainable parameters and the size of the model, the validation loss value (at convergence up to a maximum of 30 epochs of training), and the prediction interval size. With the assumption that the residuals of the predictions over the test set are normally distributed, we compute a 95\% confidence interval as 1.96 times RMSE on the validation set with the estimation resolution of 5 seconds. 
Self-attention module for decoding temporal dependencies of input waveforms improves the performance of the GATConv-EXT compared to GATConv-Ablated. The proposed GATConv-EXT outperforms the GCNConv-EXT baseline in validation loss and the confidence interval. SAGEConv-EXT model outperforms GATConv-EXT slightly, but as mentioned earlier, its aggregation mechanism is not attention-based, hence less informative in the estimation of unseen scenarios. GATConv-EXT and GATConv-INF with respectively 0.5 and 0.9 vehicles per 5-second confidence level in their estimation and a reasonably small number of parameters are our proposed architecture for attention-based inductive learning of intersection traffic flow.

\begin{table*}
    \centering
    \begin{adjustbox}{width=\textwidth,center}
        \begin{tabulary}{1.0\textwidth}{|L |L |L |L |L |L |}  
            \hline
            \hfil \textbf{Model}&
            \textbf{Parameters}&
            \textbf{Encoder Parameters}&
            \textbf{Size (MB)}&
            \textbf{Validation Loss (MSE)}& 
            \textbf{$95\%$ Confidence Interval} 
            \\
            \hline
            \textbf{GATConv-EXT} &  23.2 K &  8.2 K  & 0.093 &	0.2857 &$\mp 0.5889$\\
            \hline
            \textbf{GATConv-INF} &   50.6 K   &  8.2 K   &	 0.202    &  0.3158 &$\mp 0.8253$\\
            \hline
            \textbf{SAGEConv-EXT} &  23.0 K   &  8.1 K   &   0.092   &  0.0527 &$\mp 0.1085$\\
            \hline
            \textbf{GCNConv-EXT}  &   19.0 K 	&  4.0 K    &	 0.076    &  0.2878 &$\mp 0.5917$ \\
            \hline
            \textbf{GATConv-Ablated} &  23.0 K    &	8.1 K   &  0.092   & 0.3150   &$\mp 0.6532$\\
            \hline
    
         \end{tabulary}
     \end{adjustbox}
\renewcommand{\tablename}{Table}
\caption{\textbf{Learning performance evaluation.} The size, number of trainable parameters, validation loss value, and prediction interval size after convergence up to a maximum of 30 epochs of training on the Real-TMC dataset are compared for our proposed models (the first two rows) and the variant architectures.}
\label{table:modelpar} 
\end{table*}

Table. \ref{table:accuracy} shows the Mean Absolute Error (MAE) and Root Mean Square Error (RMSE) values at different resolutions for the estimation of exit and inflow waveforms at all lanes of all directions, collectively. The error values per indicated unit of aggregation are computed over the validation and test portions (total of 20\% split) from benchmark datasets. As shown in this table, overall, the Random-TMC dataset provides better generalization of undergoing variations in the traffic behavior, and hence better estimation of the exit/inflow waveforms. The marginal improvement at higher levels of aggregation is better for all models, which implies that estimated waveforms tend to be more accurate and consistent with the overall trends and patterns when short-term fluctuations and variations are smoothed out. In the Random-TMC dataset, as shown the SAGEConv-EXT is the only baseline that outperforms our proposed GATConv-EXT, although, as mentioned earlier, the aggregation mechanism of GAT is attention-based and can better generalize to unseen traffic scenarios. Finally, in the real-TMC scenario generation, the absolute error of exit/inflow waveform estimation by our digital twins falls below 1 vehicle unit per interval unit of 20 seconds.

For imputation, the loss function optimized over the entire feature matrix containing both stop-bar and Exit waveforms, we observed the reconstruction of the exit waveforms in the output of the model becomes hard when the model is focused on imputing only missing (Exit waveforms) in the feature matrix. We also use a dummy mask to exclude dummy lanes of an intersection in the computation of the loss values when the model is trained on a mix of different intersections each with its own specific topology. Also, we observed an increasing number of attention heads in time series decoding trade-off with the complexity of patterns and dependencies, hence its benefit becomes marginal beyond a certain number of heads. The excessive capacity of learning compared to the complexity of the problem is struggling with overfitting and may not provide substantial benefit to the performance of deep learning models. 
We used both MAE and RMSE for a comprehensive comparison of model performance. For both metrics, our proposed models outperform their variants on both datasets. 

The choice of data granularity in this study depends on the required level of detail to solve upstream signal timing optimization as described in Section \ref{appl} for any target intersection.

\begin{table*}[htbp]
    \centering
    \begin{adjustbox}{width=\textwidth,center}
        \begin{tabulary}{\textwidth}{|L |L |L |L |L |L |}

             \hline
                \multirow{2}{*}{\textbf{\hfil Aggregation}} & 
                \multicolumn{3}{l|}{\hfil\textbf{Variants}} & 
                \multirow{2}{*}{\textbf{GATConv-EXT}}&
                \multirow{2}{*}{\textbf{GATConv-INF}} \vspace{2mm}\\
                
             \cline{2-4}
             &\vspace{0.5mm}\textbf{SAGEConv-EXT}\vspace{1mm}& 
             \vspace{0.5mm}\textbf{GCNConv-EXT}\vspace{1mm}&
             \vspace{0.5mm}\textbf{GATConv-Ablated}\vspace{1mm}& 
             &\\

            \hline
            \multicolumn{6}{|l|}{\textbf{Random-TMC Dataset}}\\
            \hline
            &\hfil \textbf{MAE$|$RMSE}&\hfil \textbf{MAE$|$RMSE}&\hfil \textbf{MAE$|$RMSE}&\hfil \textbf{MAE$|$RMSE}&\hfil \textbf{MAE$|$RMSE}\\
            \hline
            5-second buckets &   0.0681$|$0.2107  &  0.2448$|$0.4447    &   0.2581$|$0.2086 &   0.2183$|$0.4466 &   0.1085$|$0.3647 \\
            \hline
            10-second buckets &   0.0936$|$0.2661  &  0.4456$|$0.7759    &   0.4748$|$0.8007  &   0.3959$|$0.7767 &  0.2053$|$0.6518   \\
            \hline
            15-second buckets &   0.1166$|$0.3115  &  0.6329$|$1.0825   &   0.6792$|$1.1188  &   0.5605$|$1.0803  &  0.2971$|$0.9171   \\
            \hline
            20-second buckets &   0.1392$|$0.3561  &  0.8099$|$1.3676    &   0.8731$|$1.4137  &   0.7210$|$1.3629 &   0.3806$|$1.1713 \\
            \hline
            \multicolumn{6}{|l|}{\textbf{Real-TMC Dataset}}\\
            \hline
            &\hfil \textbf{MAE$|$RMSE}&\hfil \textbf{MAE$|$RMSE}&\hfil \textbf{MAE$|$RMSE}&\hfil \textbf{MAE$|$RMSE}&\hfil \textbf{MAE$|$RMSE}\\
            \hline
            5-second buckets &   0.0832$|$ 0.2355 &  0.3249$|$0.5495    &   0.3502$|$0.3333  &   0.3230$|$0.3005 &   0.31970$|$0.4211 \\
            \hline
            10-second buckets &   0.1122$|$0.3012  &  0.5969$|$0.9905    &   0.6512$|$1.0420  &   0.5922$|$0.9901 &   0.6125$|$1.2084 \\
            \hline
            15-second buckets &   0.1385$|$0.3591  &  0.8507$|$1.4076    &   0.9353$|$1.4809  &   0.8430$|$1.4087 &   0.8935$|$1.7168 \\
            \hline
            20-second buckets &   0.1640$|$0.4148  &  0.8430$|$1.4087    &   1.2111$|$1.8955  &   1.0858$|$1.8044 &   1.1582$|$2.2016 \\
            \hline
        \end{tabulary}
    \end{adjustbox}
    \renewcommand{\tablename}{Table}
    \caption{\textbf{Accuracy of estimation at different aggregation resolutions.} The learning performance of and proposed digital twin and the variants are compared on our benchmark datasets in terms of aggregated mean absolute error values in vehicle-per-bucket unit of measurement.}
    \label{table:accuracy}
\end{table*}

We further investigate the latent space of Exit Time series waveforms estimation by $G_{ext}$ digital twin. We use the UMAP \cite{mcinnes1802umap} dimensional reduction algorithm for 2D visualization of the embeddings associated with each group of lanes (see Figure. \ref{fig:kde} ). Corresponding to each lane group, distinct Gaussian kernel density in the manifold learned by the model is estimated as analogous to at least 4 distinguished dynamic behaviors of traffic flow within an intersection.

\begin{figure}
        \centering
        \captionsetup{justification=raggedright,singlelinecheck=false}
        \includegraphics[width=\columnwidth]{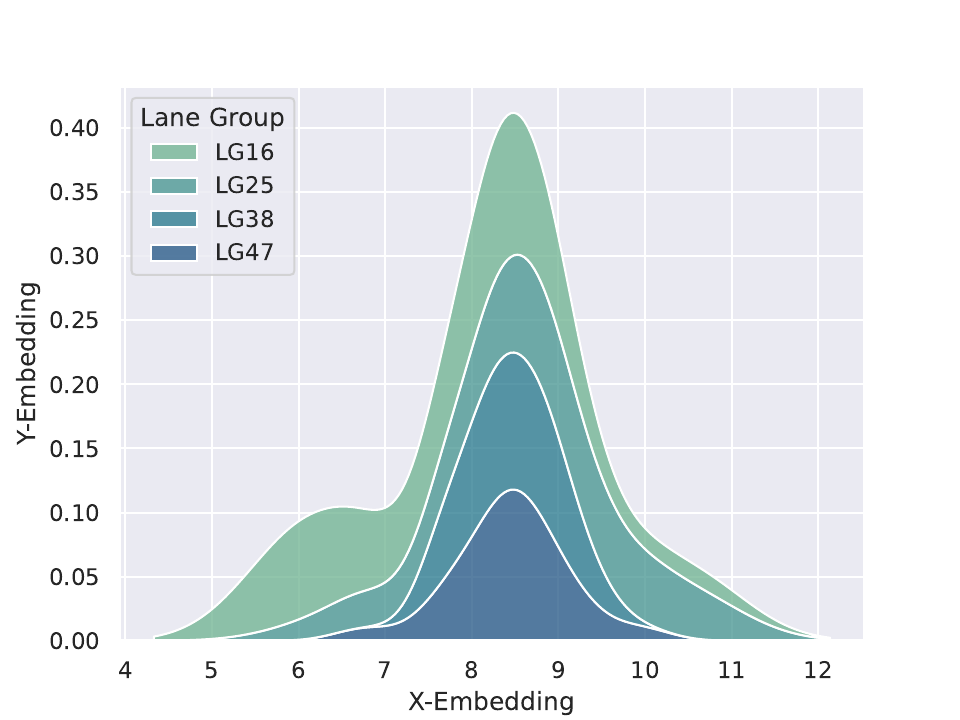}
        \renewcommand{\figurename}{Figure}
        \captionsetup{size=footnotesize}
\footnotesize
        \caption{\textbf{Kernel density estimation of traffic flow exiting from different lane groups of an intersection.} The exit time series waveform estimated by $G_{ext}$ for 1000 randomly selected samples from the test set is filtered for a specific intersection, grouped by associated lane groups (LG16: major westbound, LG25: major eastbound, LG47: minor southbound, and LG38: minor northbound) and reduced into two-dimensional embedding. The Gaussian distribution estimated for different lane groups represents four distinct dynamic behaviors of traffic flow within that intersection.}
        \label{fig:kde}
    \end{figure}

To examine the impact of different input features in the estimation of Exit Time series waveforms by $G_{ext}$ digital twin, we explain their effects using the SHAP (SHapley Additive exPlanations) values \cite{lundberg2017unified}. Because retaining Sahapley values for a graph neural network is too complex, we train a simpler explainer multivariate linear regression model mapping between input feature values and the output of $G_{ext}$ digital twin. The linear regression model can simply explain that the magnitudes of exit waveforms in the first place depend on the lane group as proof of our previous observation in Figure. \ref{fig:kde}. On average, The most effective features on the estimation of exit waveforms by the $G_{ext}$ digital twin are \textit{minimum gap}, \textit{acceleration} and \textit{cooperative lane changing} in the setting of driving behavior parameters followed by signal timing plan as a result of combining different related parameters (e.g., cycle length, barrier time, phase green times, etc., ). The complete explanation of the model performance in terms of the impact of predictive factors is shown in Figure. \ref{fig:shap}. 

\begin{figure}
        \centering
        \captionsetup{justification=raggedright,singlelinecheck=false}
        \includegraphics[width=\columnwidth]{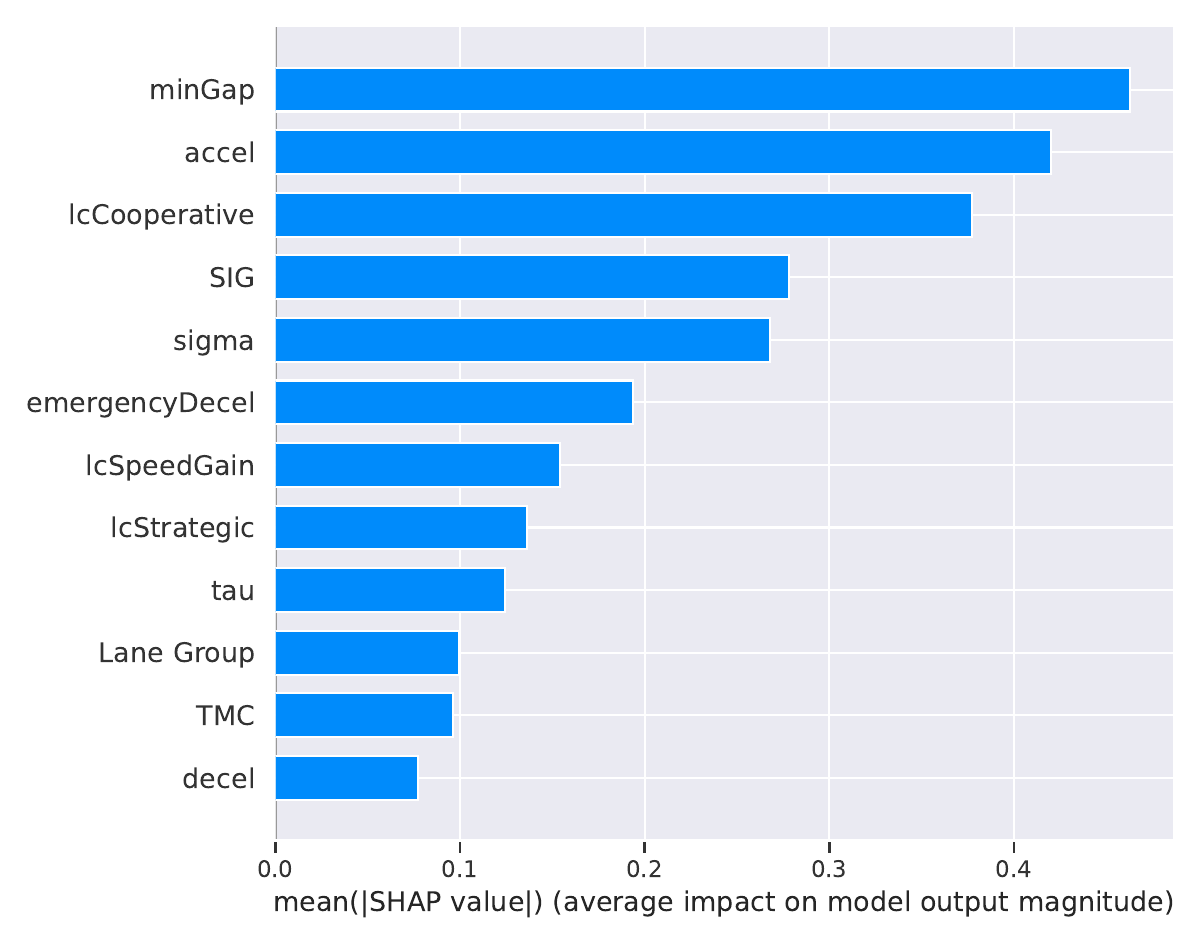}
        \renewcommand{\figurename}{Figure}
        \captionsetup{size=footnotesize}
\footnotesize
        \caption{\textbf{Average impact of features on the magnitude of estimated exit waveform.} We map the reduced dimension of the output of the model to its effective input features using a multivariate linear regression model and use it to explain our $G_{ext}$ digital twin by SHAP.}
        \label{fig:shap}
    \end{figure}

To visualize the performance of the model, we average the estimated waveforms for a batch of 1000 randomly selected samples from the test set. To simplify the visualization we only focus on a specific intersection (J3) to compare actual versus predicted lane-wise waveforms at 5-second-bucket resolution in Figure. \ref{fig:waveforms} and at 10 to 20-second-bucket resolutions in Supplementary Figures 1 to 3. There are two dummy lanes with zero actual waveforms (green colored). In higher aggregation levels, predicted and actual waveforms better match each other.


\section{Applications}
\label{appl}
This study has several practical applications, with the primary focus on Traffic Signal Timing Optimization for intersections. Given lane-wide inflow time series waveforms $inf$ estimated by graph-based digital twin $G_{inf}$ as they approach the target intersection from all intersections, we construct O-D matrices. These matrices are then used with the od2trips tool of the Simulation of Urban MObility (SUMO) software to generate route files. Each traffic scenario is evaluated using different measures of effectiveness (MOEs). The optimal signal timing plan is determined by striking the best balance among effective MOEs, driving behaviors, and safety constraints. This optimization process is facilitated by Monte Carlo Tree Search (MCTS) algorithms.

Alternatively, our digital twins can seamlessly integrate into other signal timing optimization frameworks, including those based on reinforcement learning or adaptive traffic signal control software such as TRANSYT-7F and SCOOT. These frameworks leverage techniques like Robertson platoon dispersion learning to optimize signal timing at the corridor or network level.

The outcomes of our study have broader implications for data-driven decision-making across various applications. These include traffic signal adjustments, strategic traffic planning, and infrastructure redesign for roadways and intersections. Our digital twins are particularly well-suited for traffic control design and safety management in dynamic environments, such as smart intersections.

\begin{figure*}[htp]
        \centering
        \captionsetup{justification=raggedright,singlelinecheck=false}
           \includegraphics[scale=0.15]{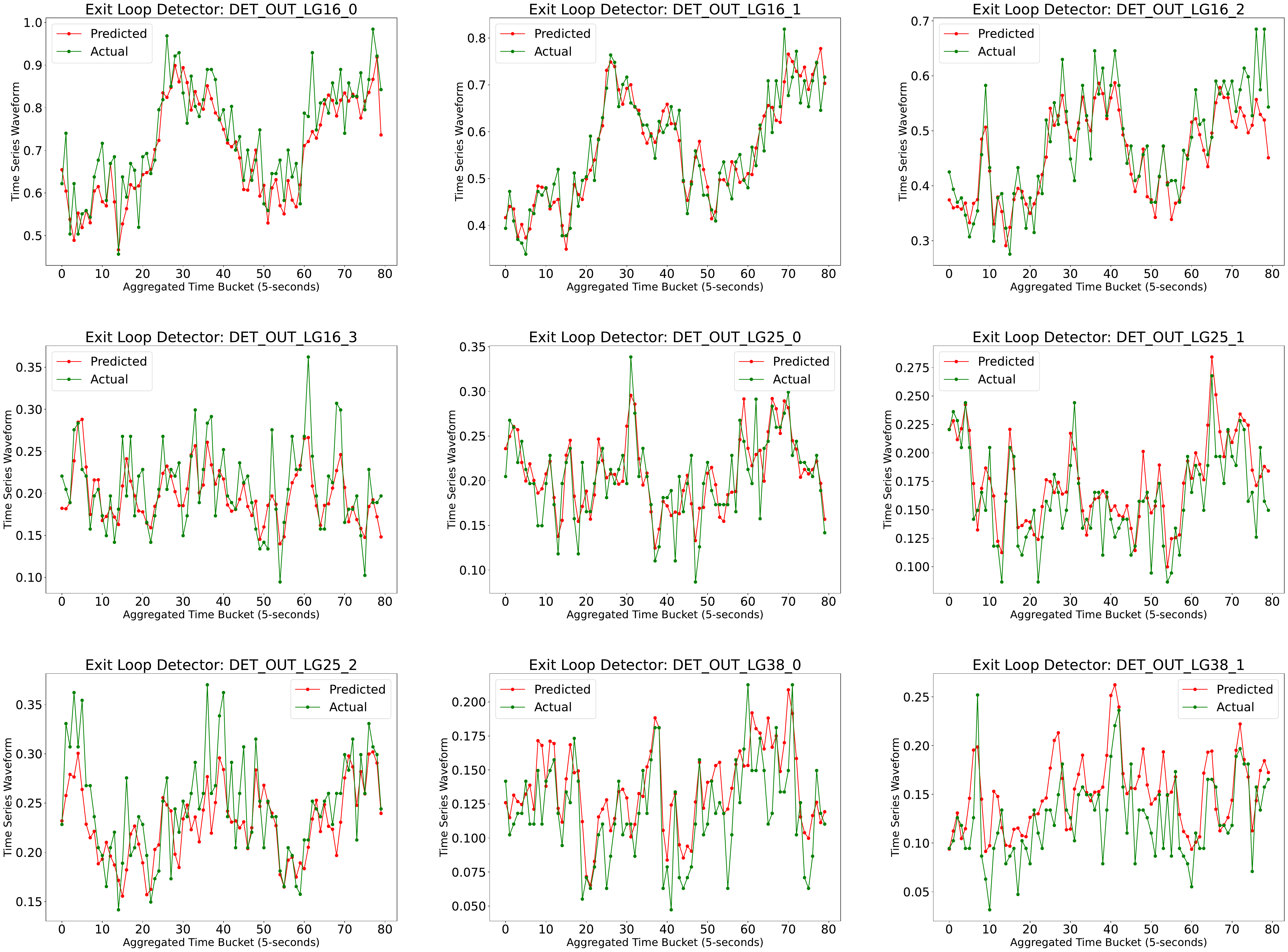}
        \renewcommand{\figurename}{Figure}
        \captionsetup{size=footnotesize}
\footnotesize
        \caption{\textbf{Comparing actual (green-colored) and predicted (red-colored) Exit time series waveforms.} For a single intersection, predicted time series waveforms reasonably match the actual ones. The same models can estimate lane-wise waveforms at all directions of different intersections. The plot shows Exit waveforms estimated at a 5-second-bucket resolution for a sample intersection (J3).}
        \label{fig:waveforms}
    \end{figure*}

\section{Related Work} 
\label{sec:related}

Existing methods of traffic state estimation can be divided into two categories model-driven and data-driven methods. Model-driven methods use macroscopic traffic models, while data-driven methods extract spatiotemporal features from historical traffic state detector data and apply statistical or machine-learning approaches to infer real-time traffic states.

 Data-driven methods were initially focused on statistical methods to predict the traffic waveforms of the same loop detector as the input. For example, deep learning with non-parametric regression was proposed by \cite{876393:20254763}, and Autoregressive Integrated Moving Average (ARIMA) was proposed by \cite{876393:20254649}. However, they are not able to model the dispersion of the platoon approaching an intersection or downstream traffic behavior while crossing the intersection. Recently, data-driven methods target traffic state estimation. For example, \cite{876393:20254909} coupled graph embedding with a generative adversarial network, and \cite{876393:20255054} combines shockwave analysis with a Bayesian Network. 

 In the literature, Machine learning approaches have been applied to traffic data for traffic state prediction \cite{876393:20253980} for different time horizons: short-term (between 5 and 30 minutes), medium-term (between 30 and 60 minutes), and long-term (over 1 hour) time windows.  

Works such as \cite{876393:20255004} and \cite{876393:20255006} use geometric deep learning architectures for to predict flow at the downstream intersection, given upstream intersection detector flows.

The seminal work in modeling inflow and outflow waveforms at traffic intersections is proposed in \cite{karnati2021subcycle}, which uses recurrent neural networks for modeling waveforms at exit, inflow, and outflow detectors. This work motivates the problem of modeling such fine-grained waveforms and their usefulness for understanding traffic flow dynamics across intersections under various signal timing plans. An extension of this work to impute measures of effectiveness (such as wait time), for a single intersection, for different signal timing plans is proposed in \cite{karnati2021intertwin}. This highlights an important application of waveform modeling using neural networks and its potential use in signal timing optimization software. However, this prior work only focuses on single intersection scenarios with simple topology (single lane with an additional left turn buffer).

Another important concern is the availability of suitable data to create generalized digital twin models, as opposed to training models to predict future time series from past data. While several induction loop detector datasets exist, they are generally unsuitable. Datasets such as the Seattle Inductive Loop Detector Dataset \cite{cui2019traffic} and Los Angeles County \cite{8809901} contain data at the 5-second resolution, which is too coarse for fine-grained sub-cycle modeling. \cite{genser2023traffic} contains loop detector data at the 1-second resolution but from only one intersection. Hence it was necessary to generate a diverse dataset, covering both realistic and unrealistic traffic scenarios and signal plans.

\section{Conclusions and Future Work}
\label{conclusion}
In this paper, we delve into the challenge of achieving comprehensive and fine-grained traffic flow estimation at urban intersections. We introduce two innovative graph neural networks so-called "Digital Twins" tailored specifically for modeling traffic flow at intersections. These digital twins can model traffic flow dynamics approaching and exiting an intersection from every single lane in any direction simultaneously. The common structure of the graph representation of intersections enables the seamless application of our digital twins to any other intersection, regardless of variable lane design and configuration.

We successfully train and evaluate our models using both real-world and completely counterfactual (randomized) traffic scenarios. The attention mechanism integrated into the graph message propagation scheme exhibits promising inductive generalization capabilities to unseen traffic patterns and traffic conditions. Traffic patterns encompass various recurring factors such as roadway design, lane configuration, traffic signal plans, and turning-movement counts, while traffic conditions include other factors effective in the current state of traffic flow including volume, density of vehicles, and driving behavior.

To systematically generate large datasets for our experiments, we leverage a Python multiprocessing package integrated with the SUMO micro-simulator. We conduct 40,000 hours of simulation span across 9 signalized intersections with varying geometries and characteristics. Each simulation scenario is recorded while randomly varying significant parameters in the configuration of signal timing plans and driving behaviors, and perturbing Origin-Destination (O-D) matrices to alternative route files within predefined ranges.

For both the Exit and Inflow waveform reconstruction experiments, we design a specific graph-based digital twin using a graph auto-encoder architecture with an imputation mechanism. The GATConv encoder layer used in our digital twins can effectively encode temporal, spatial, and contextual dimensions of traffic flowing within an intersection, exploiting various effective factors. The results demonstrate that although our proposed digital twins utilize a relatively small number of learnable parameters compared to existing deep-learning-based approaches, they perform accurately and reliably for lane-wise and fine-grained waveform reconstruction at the exit and 1-hop away road segments of any intersections.

Furthermore, we evaluate several variants of our proposed digital twins to highlight the effectiveness of the attention mechanism in traffic flow reconstruction. Additionally, to gain insights into the reconstruction mechanism learned by our digital twin, we employ some dimensionality reduction and explanation tools to visualize the manifold of hidden representations and rank the significance of input parameters in the estimation made by the models.

The results of our $G_{inf}$ digital twin can be directly applied to signal timing optimization of any standard NEMA phasing intersection, as described in Section \ref{appl}. Moreover, an additional advantage of the GATConv layer used in the architecture of our proposed digital twins is its ability to provide side information on multi-head attention coefficients automatically computed by the message propagation scheme. Further insights will be gained by investigating the relation between these attention coefficients derived by graph neural networks and recorded turning-movement counts, representing a promising future research direction.

\section{Acknowledgments}
The work was supported in part by NSF CNS 1922782. The opinions, findings, and conclusions expressed in this publication are those of the authors and not necessarily those of NSF. The authors also acknowledge the University of Florida Research Computing for providing computational resources and support that have contributed to the research results reported in this publication.


\bibliographystyle{abbrv}

\bibliography{ref}
\newpage
\section{Biography Section}
\vskip -2\baselineskip plus -1fil
\begin{IEEEbiography}[{\includegraphics[width=1in,height=1.25in,clip,keepaspectratio]{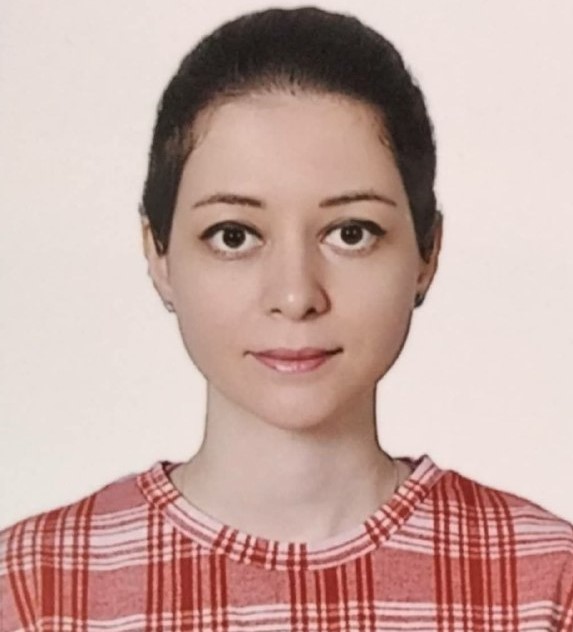}}]{Nooshin Yousefzadeh}
is currently pursuing  Ph.D. Degree with the Department of Computer and Information Science and Engineering, University of Florida, Gainesville, FL, USA. Her current research interests are in Explainable Artificial Intelligence and data-driven Machine Learning Algorithms for practical applications in Intelligent Transportation, Health Care, and Sustainability Science. 
\end{IEEEbiography}

\vskip -2\baselineskip plus -1fil
\begin{IEEEbiography}
[{\includegraphics[width=1in,height=1.25in,clip,keepaspectratio]{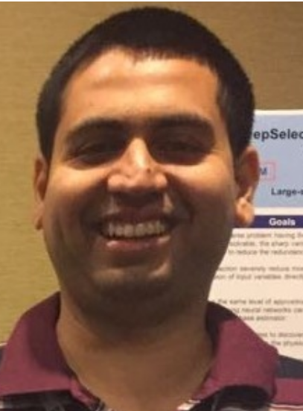}}]{Rahul Sengupta}
is a Ph.D. student at the Computer and Information Science Department at University of Florida, Gainesville, USA. His research interests include applying Machine Learning models to sequential and time series data, especially in the field of transportation engineering.
\end{IEEEbiography}

\vskip -2\baselineskip plus -1fil
\begin{IEEEbiography}[{\includegraphics[width=1in,height=1.25in,clip,keepaspectratio]{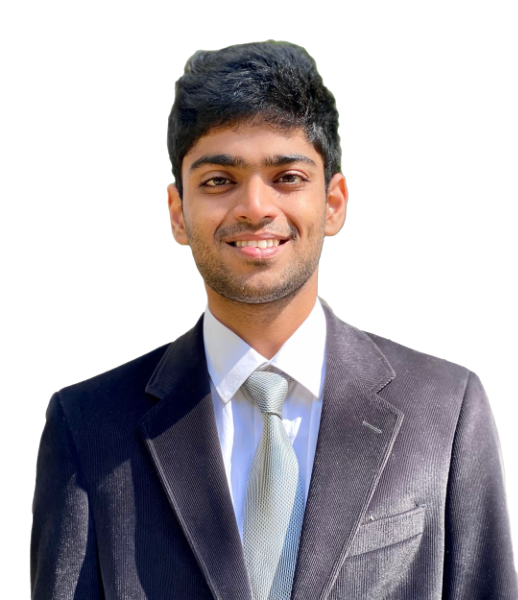}}]{Yashaswi Karnati}
completed his  Ph.D. Degree at the Department of Computer and Information Science and Engineering, University of Florida, Gainesville, FL, USA. He received a master's degree in Computer Science from University of Florida and a bachelor's degree in Electrical Engineering from Indian Institute of Technology, Dhanbad. He currently works with NVIDIA Corporation on Digital Twins. His research interest is in developing data-driven Machine Learning algorithms for practical applications in Intelligent Transportation, Health Care, and Climate Science.  
\end{IEEEbiography}

\vskip -2\baselineskip plus -1fil
\begin{IEEEbiography}[{\includegraphics[width=1in,height=1.25in,clip,keepaspectratio]{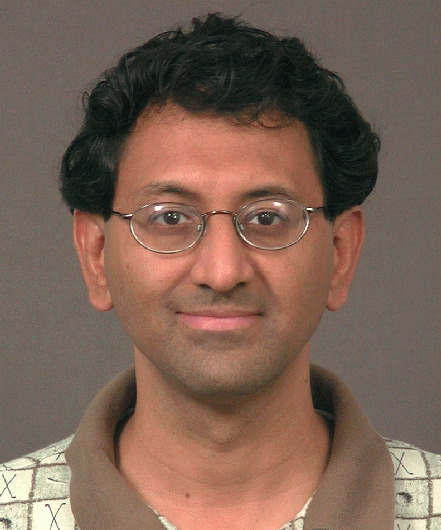}}]{Anand Rangarajan}
(Member, IEEE) is currently a Professor with the Department of Computer and Information Science and Engineering, University of Florida, Gainesville, FL, USA. His research interests include computer vision, machine learning, medical and hyperspectral imaging, and the science of consciousness
\end{IEEEbiography}

\vskip -2\baselineskip plus -1fil
\begin{IEEEbiography}[{\includegraphics[width=1in,height=1.25in,clip,keepaspectratio]{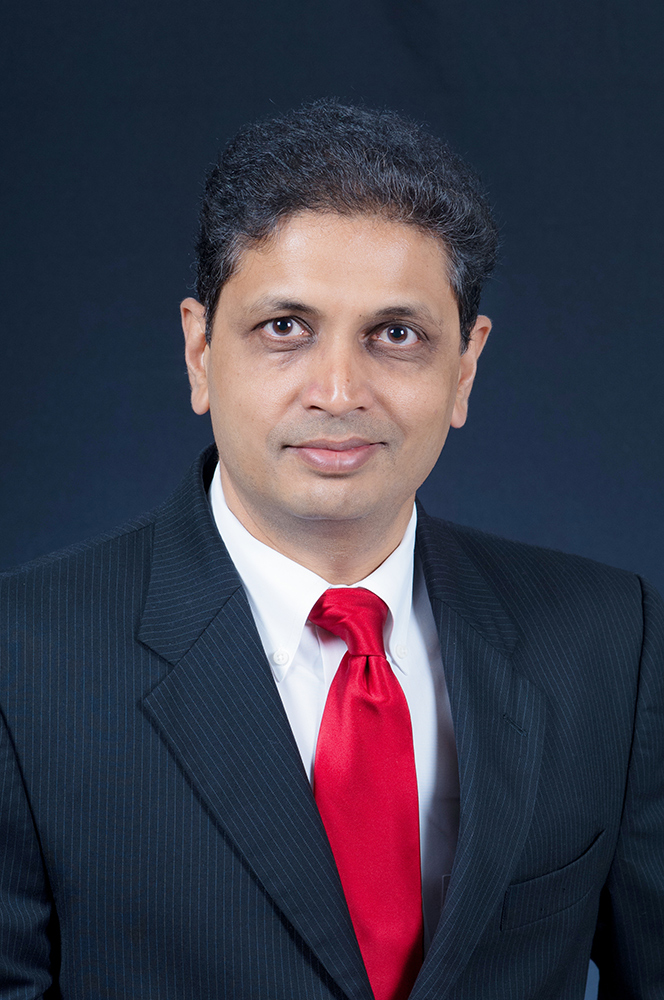}}]{Sanjay Ranka}
(Fellow, IEEE) is a Distinguished Professor in the Department of Computer Information Science and Engineering at University of Florida. His current research is on developing algorithms and software using Machine Learning, the Internet of Things, GPU Computing, and Cloud Computing for solving applications in Transportation and Health Care. He is a fellow of the IEEE, AAAS, and AIAA (Asia-Pacific Artificial Intelligence Association) and a past member of IFIP Committee on System Modeling and Optimization. He was awarded the 2020 Research Impact Award from IEEE Technical Committee on Cloud Computing. His research is currently funded by NIH, NSF, USDOT, DOE, and FDOT. From 1999-2002, as the Chief Technology Officer and co-founder of Paramark (Sunnyvale, CA), he conceptualized and developed a machine learning-based real-time optimization service called PILOT for optimizing marketing and advertising campaigns. Paramark was recognized by VentureWire/Technologic Partners as a Top 100 Internet technology company in 2001 and 2002 and was acquired in 2002.

\end{IEEEbiography}


\end{document}


\section*{Supplementary Information}

\clearpage
\section{Supplementary Tables}

\begin{table*}[htbp]
\centering



    \begin{adjustbox}{width=\textwidth,center}
    \begin{tabulary}{1.2\textwidth} {|l | L | L|}
    \hline
        \textbf{Parameter} &\hfil \textbf{ Description} &\hfil \textbf{ Variation} \vspace{2mm}\\
        \hline
        \multicolumn{3}{|l|}{\textbf{Signal Timing Plan Parameters}}\\
        \hline
        Route files &
          Nature of route files used for generating traffic &
          Of the 400,000 exemplars, around 200K are based on real-world traffic flows, and the  remaining 200K are based on randomly-generated traffic flows\\
        Cycle Lengths &
          Length of the common cycle for the corridor &
          Varies from 150 seconds to 240 seconds\\
        Offsets &
          Offsets in the start time of the cycles at various intersections &
          Offsets are varied randomly from 0 to the cycle length of the respective intersections\\
        Phase Order &
          Phase order in the dual rings of Ring-and-Barrier operation &
          Phase orders fixed for each intersection based on field settings\\
        Barrier Times &
          When the barrier in Ring-and-Barrier occurs, separating the non-coordinated phases and the coordinated phases &
          Occurs randomly while ensuring minimum allowable green times for the phases along with yellow and red times are met.\\
        Phase Lengths &
          Length of Green, Yellow, and Red times for the phases making up the dual rings of Ring-and-Barrier operation &
          Minimum and Maximum Green times, Yellow and Red times are fixed based on field settings\\
        \hline
        \multicolumn{3}{|l|}{\textbf{Driving Behavior Parameters}}\\
        \hline
        accel &
          SUMO parameter for vehicle acceleration &
          From 1.6 to 3.6 meters per second squared\\
        decel &
          SUMO parameter for vehicle deceleration &
          From 3.0 to 6.0 meters per second squared\\
        emergencyDecel &
          SUMO parameter for maximum possible deceleration for a vehicle &
          From 6.0 to 12.0 meters per second squared\\
        minGap &
          SUMO parameter for empty space left when following a vehicle &
          From 1.0 to 4.0 meters \\
        sigma &
          SUMO parameter for driver imperfection with 0 denoting perfect driving, as per SUMO's default car-following model &
          From 0.1 to 1.0\\
        tau &
          SUMO parameter for modeling a driver's desired minimum time headway &
          From 0.1 to 3.0 seconds\\
        lcStrategic &
          SUMO parameter for eagerness for performing strategic lane changing, with 0 indicating no unnecessary lane-changing &
          From  0.1 to 3.0\\
        lcCooperative &
          SUMO parameter for willingness to perform cooperative lane changing, with lower values indicating reduced cooperation &
          From  0.1 to 1.0\\
        lcSpeedGain &
          SUMO parameter for eagerness to perform lane changing to gain speed, with higher values indicating more lane-changing &
          From 0.1 to 3.0\\
        speedFactor &
          SUMO parameter for controlling an individual's speeding behavior, as a multiplier applied to the speed limit. This allows individual vehicles to overspeed based on a normal distribution &
          Normal distribution with Means ranging from 1.0 to 1.5 and Standard Deviation from 0.1 to 2.0 \\

    \hline
    \end{tabulary}
    \end{adjustbox}
\renewcommand{\tablename}{Supplementary Table}
\caption{Dataset Generation Variability}
\label{table:datagen}
\end{table*}

\clearpage
\section{Supplementary Figures}
\vskip -2\baselineskip plus -1fil
\begin{figure*}[htbp]
        \centering
        \captionsetup{justification=raggedright,singlelinecheck=false}
           \includegraphics[scale=0.15]{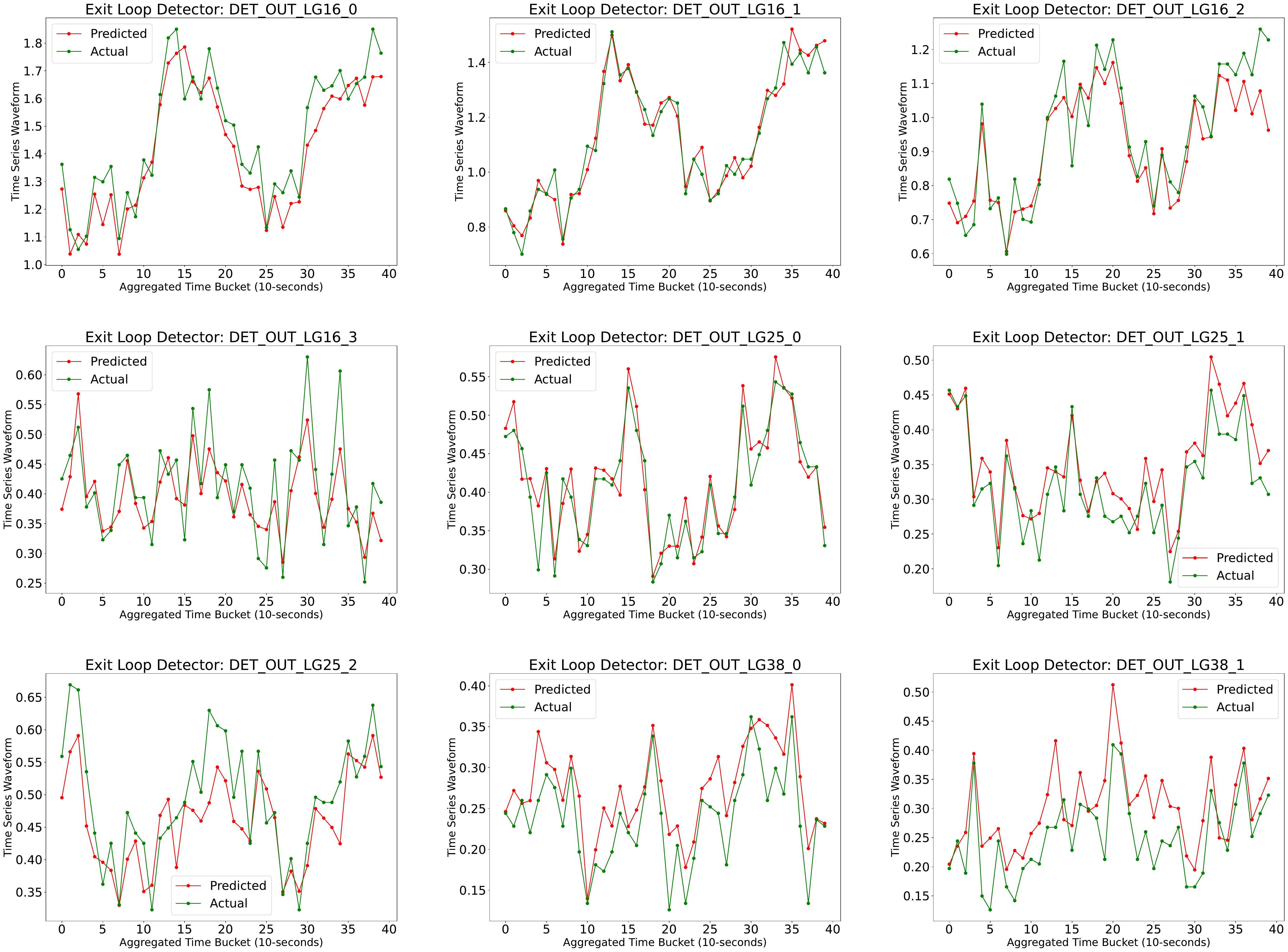}
            \renewcommand{\figurename}{Supplementary Figure}
        \captionsetup{size=footnotesize}
\footnotesize
        \caption{\textbf{Comparing actual (green-colored) and predicted (red-colored) Exit time series waveforms.} The plot shows Exit waveforms estimated at 10-second-bucket resolution for a sample intersection (J3) as an example, where lanes DET\_OUT\_LG16\_3 and DET\_OUT\_LG38\_1 are dummy lanes in its topology with respect to the generic form of an intersection topology.}
        \label{fig:waveforms}
    \end{figure*}

\vskip -2\baselineskip plus -1fil
\begin{figure*}[htbp]
        \centering
        \captionsetup{justification=raggedright,singlelinecheck=false}
           \includegraphics[scale=0.15]{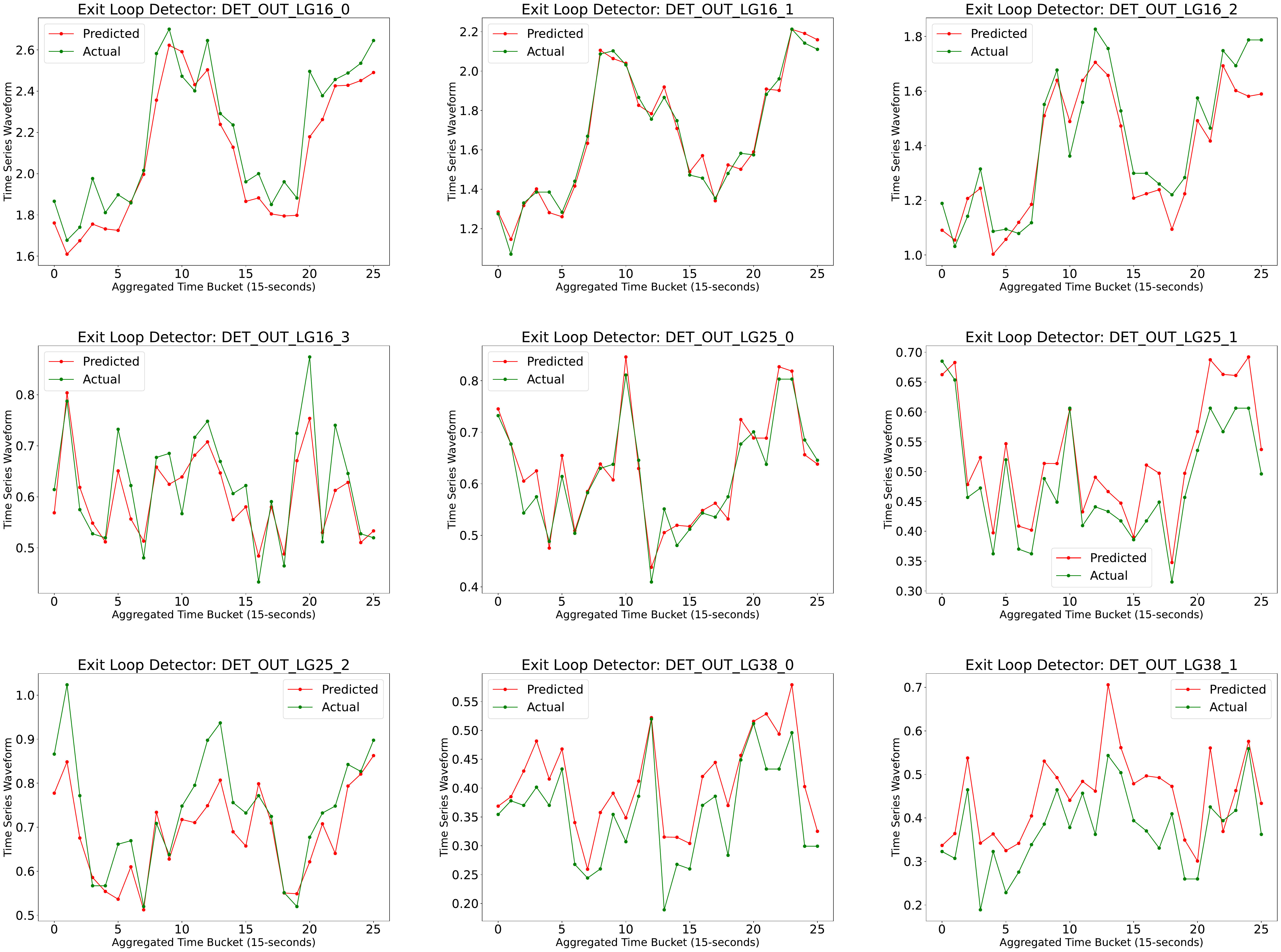}
            \renewcommand{\figurename}{Supplementary Figure}
        \captionsetup{size=footnotesize}
\footnotesize
        \caption{\textbf{Comparing actual (green-colored) and predicted (red-colored) Exit time series waveforms.} The plot shows Exit waveforms estimated at 15-second-bucket resolution for a sample intersection (J3) as an example, where lanes DET\_OUT\_LG16\_3 and DET\_OUT\_LG38\_1 are dummy lanes in its topology with respect to the generic form of an intersection topology.}
        \label{fig:waveforms}
    \end{figure*}

\vskip -2\baselineskip plus -1fil
\begin{figure*}[htbp]
        \centering
        \captionsetup{justification=raggedright,singlelinecheck=false}
           \includegraphics[scale=0.15]{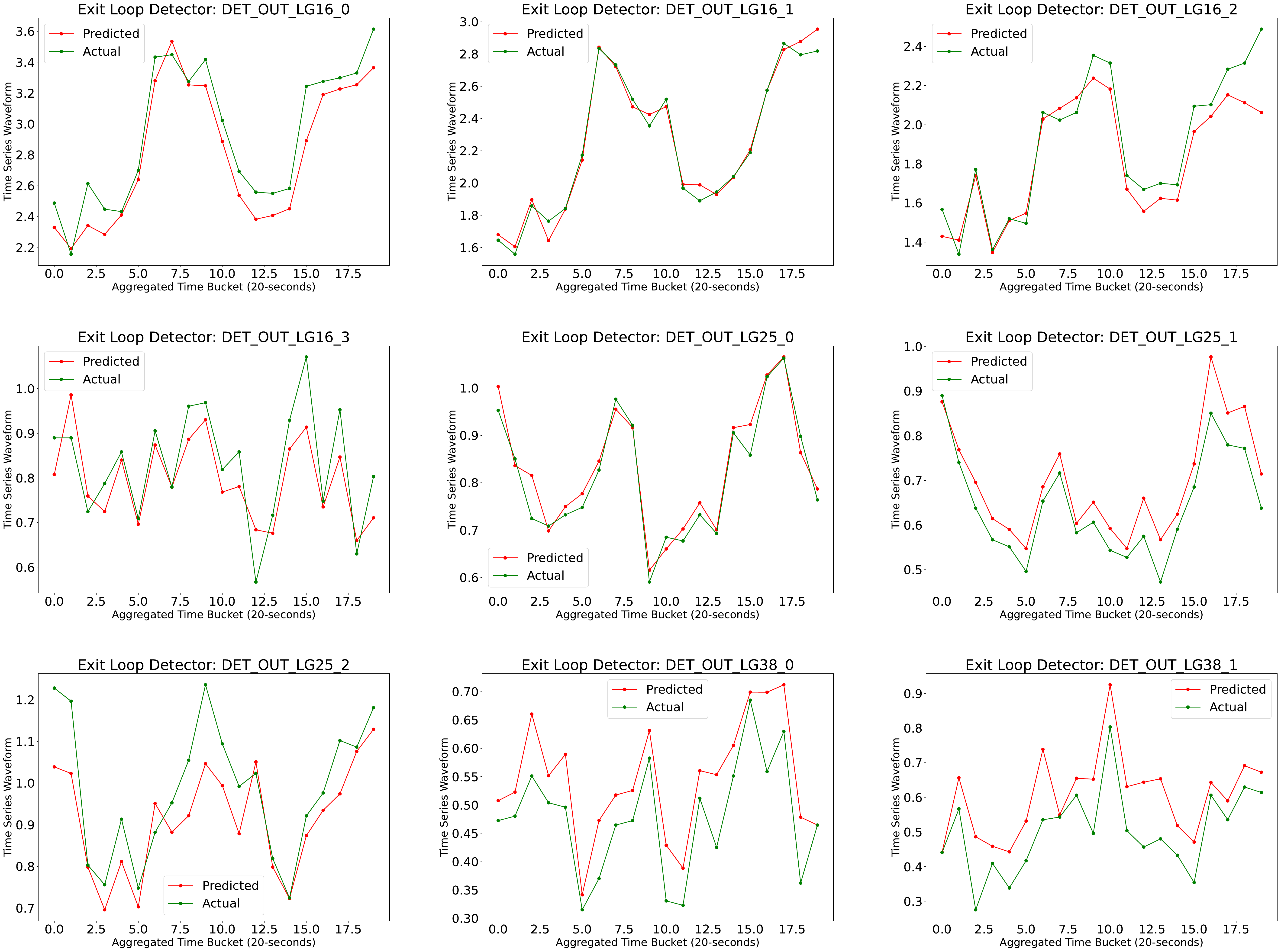}
           \renewcommand{\figurename}{Supplementary Figure}
        \captionsetup{size=footnotesize}
\footnotesize
        \caption{\textbf{Comparing actual (green-colored) and predicted (red-colored) Exit time series waveforms.} The plot shows Exit waveforms estimated at 20-second-bucket resolution for a sample intersection (J3) as an example, where lanes DET\_OUT\_LG16\_3 and DET\_OUT\_LG38\_1 are dummy lanes in its topology with respect to the generic form of an intersection topology.}
        \label{fig:waveforms}
    \end{figure*}